\documentclass[letterpaper, 10 pt, conference]{ieeeconf}  

\IEEEoverridecommandlockouts                              

\usepackage{graphicx} 
\usepackage{amsmath}
\usepackage{amsmath,amssymb}
\usepackage{booktabs} 
\usepackage{xcolor}

\title{\LARGE \bf
Multimodal Plant Root Phenotyping with Integration of 3D Skeleton Extraction and Language Analysis
}

\author{Jiakai Lin, Zijun Li, Guoyu Lu
\thanks{Jiakai Lin, Zijun Li and Guoyu Lu are with the Intelligent Vision and Sensing (IVS) Lab at Indiana University Bloomington.
        {\tt\small guoyulu62@gmail.com}}%
}

\begin{document}

\maketitle
\thispagestyle{empty}
\pagestyle{empty}

\begin{abstract}
Plant root phenotyping is fundamental to understanding below-ground structures, optimizing crop management, and improving agricultural sustainability. This paper presents a multimodal robotic AI framework that integrates 3D skeleton extraction with language-guided reasoning for interpretable and data-efficient root analysis. We develop an unsupervised skeleton extraction network based on Weighted Laplacian Contraction (W-LBC) to generate high-fidelity structural representations from dense point clouds captured by robotic 3D sensing platforms. Quantitative morphological descriptors, including root count, length, branching angle, and density, are computed from the reconstructed skeleton graph to capture geometric and topological characteristics.
Building on these features, we introduce an \textit{Evidence-First} language modeling framework that fine-tunes GPT as an interactive analytical chatbot using automatically generated instruction--response pairs. Each training sample provides measurable evidence before natural-language reasoning, enabling the model to ground interpretation in quantitative morphology. Through supervised fine-tuning, GPT associates numerical structure with semantic meaning, producing biologically consistent explanations of growth patterns and adaptive traits.
Experiments show that the structure-guided framework achieves robust, interpretable reasoning across 12 plant species with diverse root architectures. By integrating unsupervised 3D geometric perception with large-scale language understanding, our approach bridges quantitative analysis and semantic interpretation, establishing a unified paradigm for explainable robotic plant root phenotyping.
\end{abstract}

\begin{figure*}[t]
    \centering
\includegraphics[width=16cm,height=7.5cm]{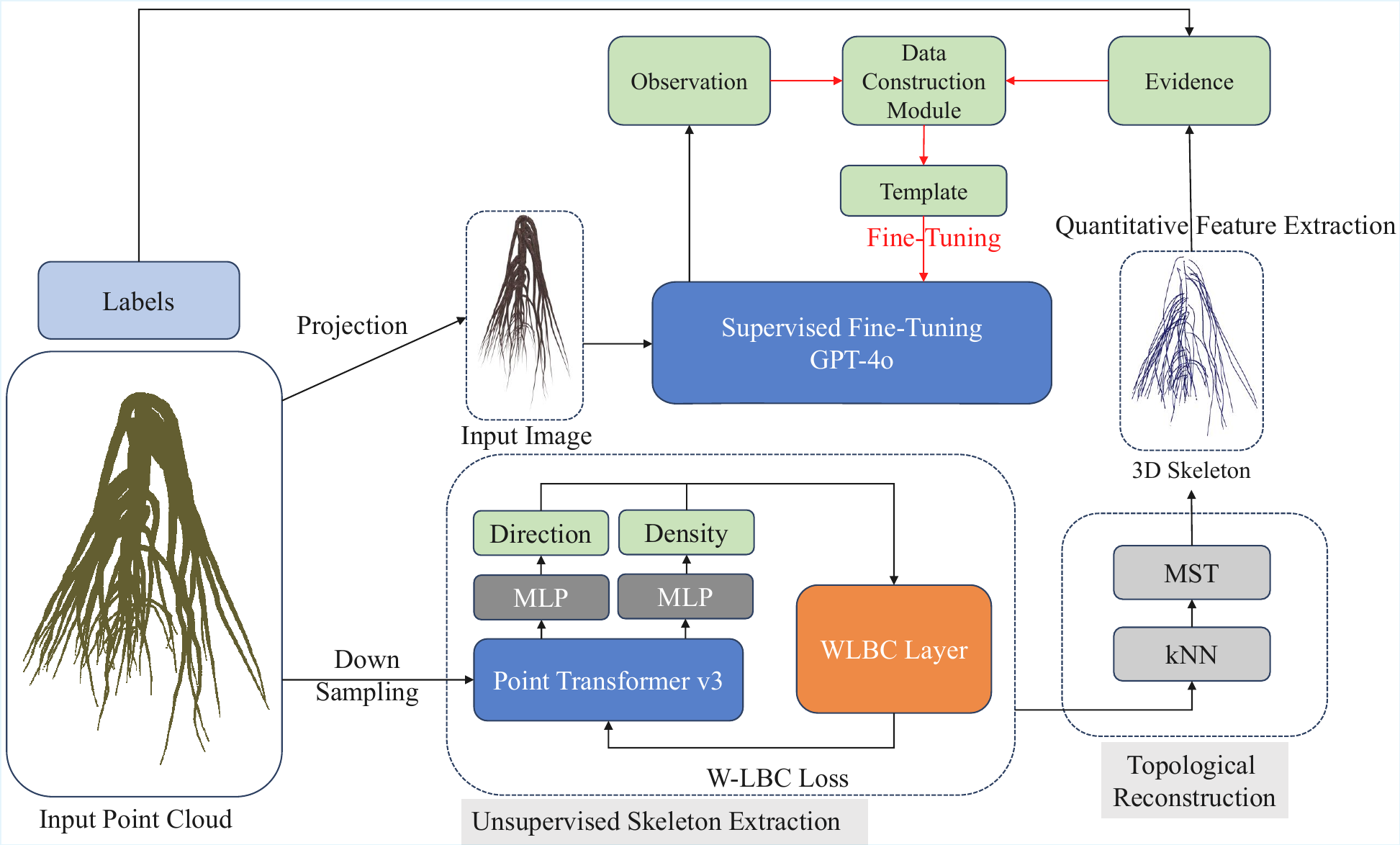}
   \vspace{-3mm}
  \caption{
Overview of the proposed multimodal framework for plant root phenotyping. 
Point clouds are processed by Point Transformer~\cite{wu2024point} with direction- and density-aware weighting under the W-LBC scheme for unsupervised 3D skeleton extraction. 
The refined skeleton is reconstructed via \(k\)-NN and MST to obtain quantitative features. 
These features and visual observations form Evidence-First templates used to fine-tune GPT-4o, enabling evidence-grounded and interpretable reasoning about root morphology.
}

    \label{fig:pdf_pipeline}
    \vspace{-6mm}
\end{figure*}

\section{Introduction}

Plant root systems are essential for water and nutrient uptake~\cite{bodner2018hyperspectral}, anchorage, and interactions with soil microbiomes~\cite{molefe2023communication}. Root structure and function directly affect plant growth, adaptability, and stress resistance~\cite{york2021phenotyping}. Yet root phenotyping remains difficult due to the low texture and structural complexity of roots and the scarcity of large, diverse, and standardized datasets~\cite{paez2015root}.

Recent studies have advanced structural representation extraction from complex objects. Wen et al.~\cite{wen2023learnable} focus on relatively simple shapes, whereas Pc\mbox{-}Skeletor~\cite{molefe2023communication} applies Laplacian contraction to produce basic root skeletons. More recent learning-based work directly extracts 3D plant root skeletons from point clouds~\cite{lin20253d}. Tree-oriented methods, including Jiang et al.~\cite{jiang2021skeleton}, AdTree~\cite{du2019adtree}, and Smart\mbox{-}Tree~\cite{dobbs2023smart}, exploit structural similarities between branches and roots. However, they often struggle with densely intertwined root systems whose topology is substantially more complex than that of tree branches.
Two major challenges remain. First, existing datasets~\cite{chang2024hyperpri,seidenthal2022iterative} contain limited root categories and insufficient multi-view, high-resolution data, hindering reconstruction in occluded and overlapping regions. Second, most approaches rely on traditional pipelines~\cite{li2022automatic} and therefore generalize poorly to unseen root structures. Fine-grained root morphology further complicates structural-property inference~\cite{akhtar2024unlocking,ju2024toporoot+}. More broadly, research in challenging robotic perception has shown the value of modality-specific sensing and representation, including subsurface radar perception~\cite{zhang2024underground,zhou2026underground}, feature extraction from minimally processed measurements~\cite{lin2025keypoint}, and adaptive event-based vision~\cite{zhang2025adaptive}. These findings underscore the need to tailor perception methods to both the sensing modality and the structural complexity of the target.
To address these challenges, we construct a multimodal robotic framework that unifies geometric perception and language reasoning for interpretable plant root phenotyping. We capture high-resolution root point clouds using a 3D scanner and develop an unsupervised skeleton extraction network based on PointTransformer~\cite{wu2024point} and Weighted Laplacian-Based Contraction (W-LBC). The proposed design introduces attention-guided point weighting to preserve fine root topology, together with geometric and linearity constraints that stabilize the contraction process and produce biologically faithful skeletons. From the reconstructed skeleton graph, quantitative descriptors, including root count, length, branching angle, and density, are computed to characterize structural variation across species.

Building on these representations, we introduce an Evidence-First framework that fine-tunes GPT \cite{cai2025tone} for reasoning grounded in measurable root morphology. Automatically generated instruction--response pairs combine numerical evidence, visual summaries, and textual explanations to infer the biological and environmental factors underlying root structures. Vision--language supervision supports semantic reasoning in geometry-related tasks~\cite{zhang2025vision}, while domain-specific adaptation improves model efficiency and specialization~\cite{codes2026}. We use structured evidence and controlled instruction templates rather than unconstrained visual prompting. Unlike systems based on fixed encoders such as CLIP or CoCa, our method incorporates domain knowledge through explicit quantitative supervision. Although evidence-grounded multimodal reasoning has shown promise in other specialized domains~\cite{yang2026interpretable}, its use in quantitative 3D root phenotyping remains limited.

We make three main contributions:
(1) a unified robotic perception-to-language pipeline connecting unsupervised 3D skeleton extraction with structure-guided language reasoning for end-to-end interpretable root analysis;
(2) an improved W-LBC framework with attention-based weighting and geometric regularization for high-fidelity root skeletonization from robotic 3D sensing; and
(3) a structure-aware GPT fine-tuning scheme that enables an evidence-grounded analytical chatbot for knowledge-based interpretation of root traits across 12 plant species.
The framework is showe in Fig.~\ref{fig:pdf_pipeline}.

\section{Related Work}
\vspace{-1mm}

\subsection{Skeletal Extraction}

Various methods have been proposed for high-quality skeleton extraction. Classical thinning on binary images, such as the Zhang--Suen algorithm, iteratively removes boundary pixels while preserving topology. Mesh contraction simplifies 3D shapes into skeletal representations. Deep approaches learn multi-task convolutional networks with multi-scale edge cues to extract object skeletons from natural images~\cite{shen2017deepskeleton}.  Recent learning-based research has further investigated skeleton detection and extraction specifically for complex 3D plant root systems~\cite{lin20253d}. Image-based 3D reconstruction has also been applied to quantitative structural and volumetric measurement in agricultural and forestry environments~\cite{lin20263d}. Nevertheless, traditional pipelines frequently degrade on complex root architectures or low-resolution observations~\cite{bohm2012methods}.
Pc\mbox{-}Skeletor relies on semantic priors, such as trunk--branch separation, to guide contraction~\cite{molefe2023communication}, whereas our approach is fully unsupervised and learns point-wise weights without external semantic labels. Pure Laplacian contraction may misconnect fine roots, collapse nearby branches, or create spurious loops in dense regions. Optimization methods in other geometric tasks have explored curvature-aware adaptation for complex trajectories~\cite{tian2026curvatureadaptiveconsistencyflowmatching}; however, such methods are not designed to preserve the topology of densely branching point clouds. Our W-LBC method instead incorporates attention-guided weighting, geometric-consistency constraints, and linearity-preservation constraints to mitigate these failures and retain fine root structures during contraction.

\vspace{-2mm}
\subsection{Visual Question Answering}
\vspace{-1mm}

Current visual question answering (VQA) research encompasses multimodal Transformers for joint image--text pretraining~\cite{chen2020uniter,tan2019lxmert,kim2021vilt}, external-knowledge augmentation~\cite{hu2023promptcap}, question-aware captioning~\cite{hu2023promptcap}, multi-answer prediction~\cite{chen2023vqa,wang2023vqa}, interpretable answer grounding~\cite{anderson2018bottom,chen2023vqa}, end-to-end architectures with reduced task-specific preprocessing~\cite{tsimpoukelli2021multimodal}, LLM-based generative VQA~\cite{brown2020language}, and task-adaptive sampling for computational efficiency~\cite{wang2023vqa}. GPT-style models have also been adapted to a wide range of visual and multimodal tasks~\cite{jia2022visual,wu2024dettoolchain,gao2025boosting,li2022blip}. Recent studies further explore domain-specific adaptation and model ensembling for small language models~\cite{codes2026}, evidence-grounded multi-agent reasoning for interpretable visual decisions~\cite{yang2026interpretable}, and vision--language knowledge for geometry-related tasks such as monocular depth estimation~\cite{zhang2025vision}. The sensitivity of model outputs to prompt tone and formulation~\cite{cai2025tone} also underscores the importance of controlled instructions in domain-specific evaluation.
CLIP-based representations show strong generalization in VQA and related multimodal tasks~\cite{sun2024alpha,zhang2024long,shen2021much}; however, most existing work remains centered on 2D visual recognition~\cite{zhang2022pointclip,huang2023clip2point}. Language-based reasoning over explicit 3D plant root structures is still underexplored. We therefore develop a domain-specific VQA system based on a plant root skeleton dataset, using measurable structural descriptors as explicit evidence to improve the accuracy and interpretability of root-structure analysis.

\section{3D Skeleton and Quantitative Analysis}

\subsection{Unsupervised 3D Skeleton Extraction}
\vspace{-1mm}
Our framework leverages PointTransformer v3~\cite{wu2024point} to extract local and global geometric features that guide a Direction--Density-Aware Weighted Laplacian-Based Contraction (W-LBC) for unsupervised 3D root skeleton extraction, as illustrated in Fig.~\ref{fig:pdf_pipeline}. 
We begin by acquiring high-resolution point clouds \(\mathbf{X} \in \mathbb{R}^{N \times 3}\) using a 3D scanner. 
The dense point cloud is first downsampled to reduce computational complexity. 
The downsampled point set is then fed into the PointTransformer v3 network, which captures attention-based contextual features for each point. 
At the output layer, two lightweight MLP heads are introduced to predict a directional vector \(\mathbf{t}_i \in \mathbb{R}^3\) and a local density index \(\rho_i \in \mathbb{R}_+\) for each point \(i\). 
These two quantities serve as geometry-aware indicators that modulate the strength and anisotropy of the contraction field. 
After obtaining the predicted \(\mathbf{t}_i\) and \(\rho_i\), the Direction--Density-Aware W-LBC process iteratively shrinks the point cloud toward its intrinsic one-dimensional skeleton while preserving topological connectivity. 
The weighted Laplacian matrix \(\mathbf{L}\) is constructed using distinct weight matrices for attraction (\(\mathbf{W}_a\)) and contraction (\(\mathbf{W}_c\)) as follows:
\vspace{-1mm}
\begin{equation}
\mathbf{L} = \mathbf{D}_a - \mathbf{W}_a + \mu (\mathbf{D}_c - \mathbf{W}_c),
\vspace{-2mm}
\end{equation}
where \(\mathbf{D}_a\) and \(\mathbf{D}_c\) are diagonal degree matrices:
\vspace{-2mm}
\begin{equation}
D_a(i,i) = \sum_{j \in \mathcal{N}(i)} W_{a,ij}, \quad D_c(i,i) = \sum_{j \in \mathcal{N}(i)} W_{c,ij}.
\vspace{-2mm}
\end{equation}

The attraction and contraction weights now incorporate both directional similarity and local density:
\vspace{-1mm}
\begin{equation}
W_{a,ij} = 
\exp\!\left(-\frac{\|\mathbf{x}_i - \mathbf{x}_j\|^2}{\sigma_a^2}\right)
\cdot 
\frac{\rho_i + \rho_j}{2}
\cdot
\exp\!\left(\kappa\, |\mathbf{t}_i^\top \mathbf{t}_j|\right),
\vspace{-1mm}
\end{equation}
\vspace{-1mm}
\begin{equation}
W_{c,ij} = 
\exp\!\left(-\frac{\|\mathbf{x}_i - \mathbf{x}_j\|^2}{\sigma_c^2}\right)
\cdot 
\frac{\rho_i + \rho_j}{2}.
\vspace{-1mm}
\end{equation}

\noindent
Here, \(\mathbf{t}_i^\top \mathbf{t}_j\) measures the local directional consistency between neighboring points, 
and \(\rho_i\) adjusts the contraction magnitude according to local density. 
The parameter \(\kappa\) controls the sensitivity to directional alignment, 
while \(\sigma_a\) and \(\sigma_c\) define the spatial influence radius for attraction and contraction, respectively. 
The iterative contraction updates the position of each point as:
\vspace{-1mm}
\begin{equation}
    \resizebox{.80\linewidth}{!}{$
        \begin{aligned}
            \mathbf{x}_i^{(t+1)} &= \mathbf{x}_i^{(t)} 
            - \eta \sum_{j \in \mathcal{N}(i)} W_{a,ij} (\mathbf{x}_i^{(t)} - \mathbf{x}_j^{(t)}) \\
            &\quad + \nu \sum_{j \in \mathcal{N}(i)} W_{c,ij} (\mathbf{x}_j^{(t)} - \mathbf{x}_i^{(t)}),
        \end{aligned}
    $}
    \vspace{-2mm}
\end{equation}

\noindent
where \(\eta\) denotes the learning rate, 
\(\mu\) is a contraction weighting factor within the Laplacian matrix, 
\(\nu\) is a contraction coefficient for spatial updates, 
and \(t\) denotes the iteration index. 
This direction--density-aware formulation enables the contraction to proceed preferentially along coherent structural directions, 
preserving root connectivity and preventing over-shrinkage in dense branching regions. 
As a result, the process yields a smooth, topologically consistent skeleton representation of the input root system.

\vspace{-1mm}
\subsection{Weighted and Geometric Constraints}
\vspace{-1mm}
The convergence of the contraction process is governed by a composite optimization objective that balances contraction strength, smoothness, and connectivity. 
The iteration stops when the total energy variation between successive steps satisfies:
$
\left|\mathcal{L}_{\text{total}}^{(t+1)} - \mathcal{L}_{\text{total}}^{(t)}\right| < \epsilon,
$
where \(\epsilon\) is a predefined threshold for convergence, and \(\mathcal{L}_{\text{total}}\) is the overall loss defined below.

\vspace{1mm}
\noindent\textbf{(1) Weighted Laplacian Contraction Loss.}
The primary contraction loss is defined as:
\vspace{-2mm}
\begin{equation}
\mathcal{L}_{\text{W-LBC}} = \|\mathbf{S} - \mathbf{X}\|_F^2 
+ \alpha_1 \, \text{Tr}(\mathbf{S}^\top \mathbf{L}_w \mathbf{S}) 
+ \alpha_2 \, \text{Tr}(\mathbf{S}^\top \mathbf{L}_w^\top \mathbf{L}_w \mathbf{S}),
\vspace{-2mm}
\end{equation}
\noindent
where \(\mathbf{S}\) is the contracted skeleton point matrix, \(\mathbf{X}\) is the input point cloud, and \(\mathbf{L}_w\) is the weighted Laplacian constructed using the direction--density-aware weights defined in Eq.~(7)--(8). 
The first term keeps the contracted points close to the original geometry, 
while the trace-based Laplacian regularizers maintain local smoothness and prevent over-contraction in high-density regions. 
The coefficients \(\alpha_1\) and \(\alpha_2\) control the balance between structural preservation and contraction strength.

\vspace{1mm}
\noindent\textbf{(2) Directional Smoothness Loss.}
To encourage directional continuity along root branches, we define a smoothness constraint on the predicted direction vectors:
\vspace{-2mm}
\begin{equation}
\mathcal{L}_{\text{smooth}} = 
\frac{1}{N} \sum_{i=1}^{N} 
\left\| \mathbf{t}_i - \frac{1}{|\mathcal{N}(i)|}
\sum_{j \in \mathcal{N}(i)} \mathbf{t}_j \right\|^2.
\vspace{-2mm}
\end{equation}
\noindent
This term penalizes abrupt directional changes between neighboring points, enforcing spatially coherent flow along root axes and reducing local jittering during contraction.

\vspace{1mm}
\noindent\textbf{(3) Connectivity Preservation Loss.}
To maintain topological integrity and prevent fragmentation of thin root branches, we further introduce a connectivity term based on local density similarity:
\vspace{-2mm}
\begin{equation}
\mathcal{L}_{\text{conn}} =
\sum_{(i,j) \in \mathcal{E}}
|\rho_i - \rho_j| \,
\exp\!\left(-\frac{\|\mathbf{x}_i - \mathbf{x}_j\|^2}{\sigma^2}\right),
\vspace{-2mm}
\end{equation}
\noindent
where \(\mathcal{E}\) denotes neighboring point pairs, and \(\rho_i, \rho_j\) are the predicted local density indices. 
This term enforces consistency between adjacent points with similar density, 
preserving continuous structures across varying local sampling densities.

\vspace{1mm}
\noindent\textbf{(4) Overall Objective.}
The total optimization objective integrates the above components as:
\vspace{-1mm}
$
\mathcal{L}_{\text{total}} = 
\mathcal{L}_{\text{W-LBC}} 
+ \beta_1 \mathcal{L}_{\text{smooth}} 
+ \beta_2 \mathcal{L}_{\text{conn}},
$
\noindent
where \(\beta_1\) and \(\beta_2\) are weighting coefficients balancing smoothness and connectivity constraints.
This unsupervised objective allows the network to refine the skeleton structure using learned geometric priors, 
without requiring any ground-truth annotations.
The resulting contraction process yields topologically faithful, directionally aligned skeletons that accurately represent the fine-scale morphology of plant root systems.

\subsection{Topological Reconstruction and Quantitative Feature Extraction}

After the contraction process, the refined skeleton points \(\mathbf{S}=\{\mathbf{s}_i\}_{i=1}^N\) are converted into a graph representation for quantitative morphological analysis. 
Rather than relying solely on a Minimum Spanning Tree (MST), we employ a multi-stage reconstruction strategy that ensures both geometric fidelity and topological completeness. 
First, a \(k\)-nearest neighbor graph \(\mathcal{G}_0=(\mathcal{V},\mathcal{E}_0)\) is constructed over \(\mathbf{S}\) to preserve local geometric adjacency, and edges longer than a threshold \(\tau_d\) are removed to prevent false long-range connections. 
From this pruned graph, a preliminary MST \(\mathcal{T}\) is extracted to establish global connectivity of the root structure.
To further refine the topology, the MST is enhanced by reconnecting fragmented endpoints and merging locally consistent loops. 
A direction-aware criterion restores small gaps without introducing spurious cycles:
\[
\mathcal{E}_{\text{refined}} = 
\{(i,j)\mid \|\mathbf{s}_i-\mathbf{s}_j\|<\tau_c,\; |\mathbf{t}_i^\top \mathbf{t}_j|>\tau_t\},
\]
where \(\tau_c\) and \(\tau_t\) denote spatial and directional thresholds, respectively. 
The final graph is given by \(\mathcal{G}=(\mathcal{V}, \mathcal{E}_{\text{MST}}\cup\mathcal{E}_{\text{refined}})\), providing a compact yet topologically faithful representation of the root skeleton.
Based on this refined topology, we compute several morphological descriptors to quantitatively characterize the root system. 
The total number of root branches \(n_r\) is obtained by counting leaf nodes, excluding the stem node:  
\(n_r = \sum_{v\in\mathcal{V}}\mathbf{1}_{\{\deg(v)=1\}} - 1.\)
The total root length \(l\) is the sum of Euclidean distances along all edges:  
\(l = \sum_{(i,j)\in\mathcal{E}}\|\mathbf{s}_i-\mathbf{s}_j\|\).  
The hierarchical depth \(d_h\), representing the longest growth path, is defined as  
\(d_h = \max_{P\subseteq\mathcal{G}} \sum_{(i,j)\in P}\|\mathbf{s}_i-\mathbf{s}_j\|\).  
The mean branching angle \(\bar{\theta}_b\) quantifies the average divergence at bifurcations:  
\(\bar{\theta}_b = \frac{1}{|\mathcal{E}_b|}\sum_{(i,j)\in\mathcal{E}_b}\arccos(|\mathbf{t}_i^\top\mathbf{t}_j|)\),  
where \(\mathcal{E}_b\) denotes the set of edges connected to branching nodes.  
Finally, the root density \(\rho_r\) measures structural compactness as the ratio between total root length and the skeleton’s bounding volume \(V_{\text{bbox}}(\mathbf{S})\):  
\(\rho_r = l / V_{\text{bbox}}(\mathbf{S})\).

\section{Multimodal Reasoning and Language-Guided Root Analysis}

The previous chapter introduced a geometry-driven framework for unsupervised 3D root skeleton extraction and quantitative feature computation.
While descriptors such as root count, total length, and branch density capture structural variation, they cannot alone reveal biological or environmental implications.
Researchers interpret such patterns in context: dense branching may indicate adaptation to compact soil, whereas elongated roots may suggest nutrient limitation.
Motivated by this analogy, we bridge quantitative morphology and language reasoning by enabling large language models (LLMs) to interpret root structures through explicit evidence-grounded prompts.
Traditional vision--language models describe visual content but lack structural awareness.
In contrast, GPT-based multimodal models possess world knowledge and reasoning ability but require structured inputs to ground responses.
To address this gap, we design a structure-guided multimodal framework centered on an \textit{Evidence-First Output Format} that binds quantitative and topological features to language.
This chapter presents dataset construction and the fine-tuning procedure aligning GPT’s reasoning with morphological evidence.
\subsection{Data Construction and Template Design}
\vspace{-1mm}
The multimodal training data are derived from three aligned sources:
(1) original root images captured by the 3D scanner,
(2) skeletonized representations from the W-LBC contraction process,
and (3) quantitative descriptors such as root count, length, mean branch angle, and density computed from the reconstructed graph.
Each sample corresponds to a unique plant instance and includes metadata labels such as species type, growth condition, and scanning context.
Although the final fine-tuning samples are textual, each instance remains linked to its images, including the raw root photograph and skeleton rendering.
During dataset construction, GPT-4o processes these images to generate the \texttt{[Observation]} field, summarizing traits such as branching density and curvature.
This design preserves the quantitative and visual grounding required for multimodal semantic alignment.
Each instruction-response pair follows a unified format with four fields:
\textbf{Evidence:} Quantitative morphological descriptors obtained from the geometric analysis pipeline.
 \textbf{Observation:} A concise summary of the root's appearance and structure, generated through rule-based parsing and refined by GPT’s editing.
 \textbf{Question:} A domain-relevant query prompting inference or explanation of biological or environmental factors.
 \textbf{Answer:} A response that explicitly cites evidence and provides higher-level reasoning.

\begin{figure}[t]
    \centering
    \includegraphics[width=0.48\textwidth]{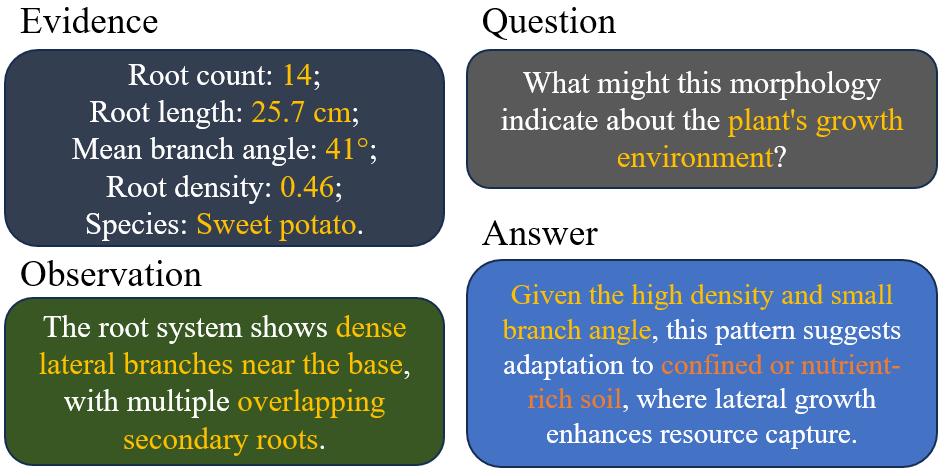}
    \vspace{-7mm}
    \caption{
    Example of the Evidence-First template for fine-tuning, where quantitative evidence, visual observation, and reasoning are explicitly linked through structured question–answer pairs.
    }
    \label{fig:evidence_template}
    \vspace{-7mm}
\end{figure}

An example of the Evidence-First template is shown in Fig.~\ref{fig:evidence_template}.
The Evidence field is populated from geometric outputs, while the Observation field derives from GPT’s visual captioning.
Automatic filtering enforces consistency between cited evidence and textual conclusions, and GPT standardizes domain-specific terminology (e.g., ``lateral expansion,'' ``branching density'') for linguistic coherence.
Through this pipeline, we achieve large-scale generation of high-quality instruction--response pairs without manual annotation.
The resulting dataset spans diverse plant species, growth conditions, and morphology patterns, providing rich supervision that tightly couples geometric structure with language reasoning, forming the basis for subsequent fine-tuning and multimodal interpretation.

\begin{figure*}[h]
    \centering
    \includegraphics[width=0.88\linewidth]{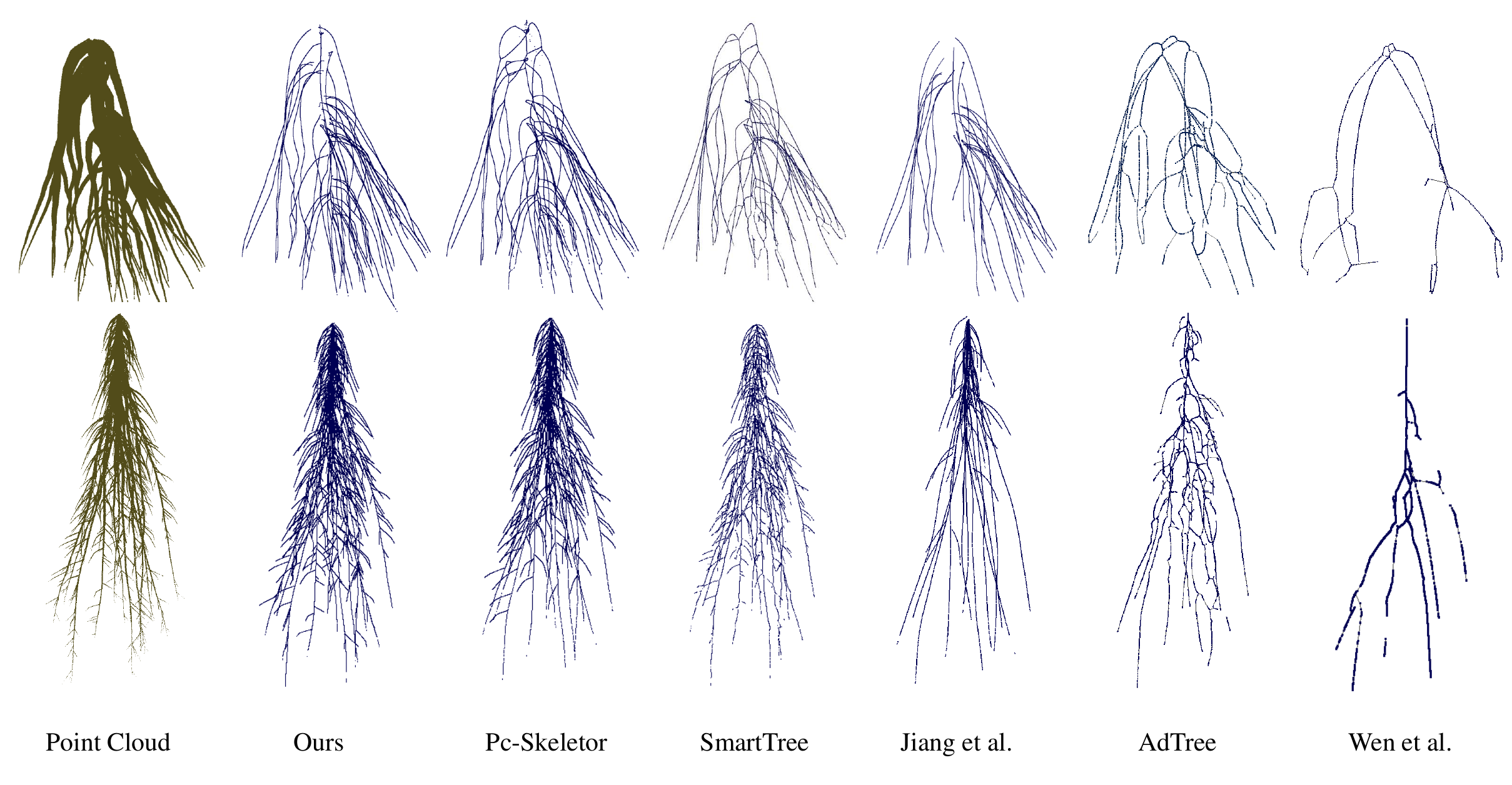}
    \vspace{-6mm}
    \caption{Comparison of root skeletons generated from input point cloud using various methods. The point cloud input is shown on the left, followed by skeletons generated by our method, Pc-Skeletor \protect\cite{molefe2023communication}, SmartTree \protect\cite{dobbs2023smart}, Jiang et al. \protect\cite{jiang2021skeleton}, AdTree \protect\cite{du2019adtree}, and Wen et al. \protect\cite{wen2023learnable}.}
\vspace{-7mm}
    \label{fig:pdf_root}
\end{figure*}
\vspace{2mm}

\subsection{Fine-Tuning Procedure}

We adopt standard \textbf{Supervised Fine-Tuning (SFT)} to align GPT's reasoning behavior with the proposed structure-aware dataset.
Unlike prompt engineering or zero-shot prompting, SFT explicitly teaches the model to produce evidence-grounded and biologically consistent responses given structured inputs.

During fine-tuning, each sample consists of an input prompt and a corresponding target completion.
The prompt contains the concatenated \texttt{[Evidence]}, \texttt{[Observation]}, and \texttt{[Question]} fields, while the target completion corresponds to the \texttt{[Answer]}.
The model is trained to maximize the likelihood of generating the target sequence conditioned on the prompt.
The training objective follows the standard cross-entropy formulation:
\vspace{-1mm}
\begin{equation}
\mathcal{L}_{\text{SFT}} = -\frac{1}{L} \sum_{t=1}^{L} \log P(w_t \mid w_{1:t-1}, \text{Prompt}),
\vspace{-1mm}
\end{equation}
where \(L\) denotes the number of tokens in the target answer.
This objective encourages the model to maintain factual consistency and linguistic coherence when responding to structured inputs.
To ensure training stability, the learning rate is conservatively set to prevent overfitting to the limited morphological domain.
We fine-tune the model with approximately 5,000 instruction--response pairs covering different species and morphology combinations.
Although the fine-tuning is lightweight, it significantly improves the model’s ability to cite quantitative evidence and maintain logical consistency across reasoning steps.

A key feature of our design is that it does not modify the underlying GPT architecture or pretraining objectives.
Instead, the model is conditioned through structured supervision to learn causal links between numerical descriptors and biological interpretations.
Empirically, this conditioning yields answers that are more interpretable and grounded.
For instance, rather than describing a sample as ``dense and healthy,'' the fine-tuned model explicitly relates its assessment to evidence fields, e.g.,
\textit{“Given the high root count (14) and small mean angle (41°), the morphology indicates lateral growth adaptation to dense soil.”}

Overall, the fine-tuning procedure transforms GPT from a general descriptive model into a reasoning-oriented agent capable of grounding its language generation on morphological evidence.
This approach bridges quantitative perception and semantic interpretation, establishing a pathway for integrating structural analysis with language-based reasoning in automated plant phenotyping.

\section{Experiment}
\vspace{-1mm}
\subsection{Dataset}
\vspace{-1mm}
\begin{figure*}[h]
    \centering
    \includegraphics[width=0.99\textwidth]{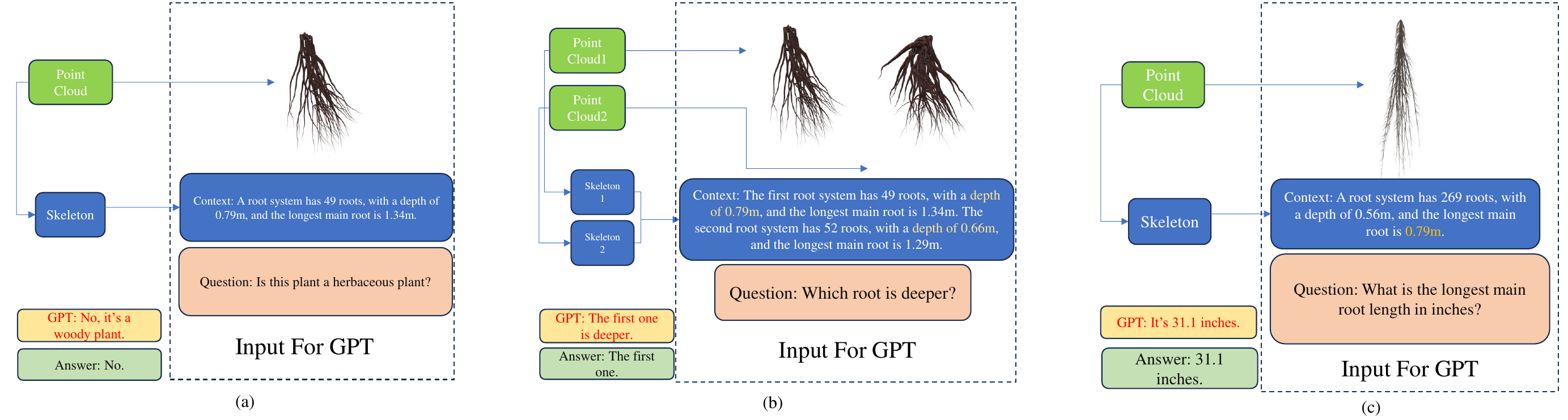}
    \vspace{-4mm}
    \caption{Our adaptive model can handle various types of test questions, with the image illustrating three example types and the corresponding responses from the adaptive GPT-4o model: (a) judgment or selection, (b) simultaneous multiple inputs, and (c) inference. }
    \label{fig:pdf_image}
    \vspace{-1mm}
\end{figure*}
Due to the lack of public datasets containing 3D root point clouds with phenotyping annotations, we created a custom dataset. We capture high-resolution 3D root point clouds and perform manual denoising before annotation. The dataset spans 12 species, including two primary species and ten complementary species with 1--5 samples each, covering simple and highly entangled root architectures. It contains 400 Sweet Potato root models with 200--400 roots each and 800 Apple Tree root models with 20--60 roots each. Each model was manually annotated with key phenotypic traits, including visible root count and longest root length, along with metadata such as species, growth stage, and collection location. To enhance diversity and generalizability, we added ten additional species, including simpler structures such as Grass and Onion and complex ones such as Peanuts and Sorghum, with 1--5 samples per species.
The dataset was split into training, validation, and testing sets in a 70-20-10 ratio, ensuring balanced representation across plant types and root structures. The same split was used for both the 3D skeleton extraction model and the fine-tuned GPT-4o model, preventing data leakage and ensuring fair evaluation of the integrated framework.

\textbf{Implementation Details.}
Our framework integrates two complementary components: a geometry-driven 3D skeleton extraction module and a language-guided reasoning module for phenotypic interpretation. Both are trained independently to ensure geometric precision and semantic consistency.
The skeleton extraction network adopts the PointTransformer backbone from Chapter~3, enhanced with direction- and density-aware weighting to guide the Weighted Laplacian-Based Contraction (W-LBC). Training is unsupervised for 50 epochs with a batch size of 4 and an initial learning rate of 0.001, decayed by 0.5 every 5 epochs. The combined contraction, geometric consistency, and linearity losses ensure smooth and topology-preserving skeleton convergence.
The reasoning module fine-tunes GPT-4o via supervised fine-tuning (SFT) on the structured Evidence-First dataset described in Chapter~4. Each sample contains quantitative descriptors, a visual observation from skeleton renderings, and a paired question–answer. Fine-tuning through OpenAI’s API follows the standard cross-entropy objective, guiding the model to produce evidence-grounded and biologically coherent responses. Rather than relying only on visual alignment, the fine-tuned GPT-4o explicitly reasons over numerical evidence and contextual cues.
All experiments use the same train–test split. Non-learning baselines (e.g., AdTree, Jiang \textit{et al.}) are evaluated directly, while all learning-based methods share identical hyperparameters. Reasoning evaluation is performed on held-out structured prompts, measuring evidence consistency, reasoning quality, and linguistic coherence.

\begin{table}[t]
\vspace{-3mm}
    \centering
    \resizebox{0.48\textwidth}{!}{
    \begin{tabular}{lcccc}
        \toprule
        \textbf{Model} & \textbf{RCA} & \textbf{LRA} & \textbf{RCA} & \textbf{LRA} \\
                      & \textbf{(\textless{}100)} & \textbf{(\textless{}100)} & \textbf{(\textless{}500)} & \textbf{(\textless{}500)} \\
        \midrule
        Wen et al. \protect\cite{wen2023learnable} & 0.22 & 0.56 & 0.15 & 0.49 \\
        AdTree \protect\cite{du2019adtree} & 0.45 & 0.67 & 0.33 & 0.62 \\
        Jiang et al. \protect\cite{jiang2021skeleton} & 0.53 & 0.71 & 0.48 & 0.63 \\
        SmartTree \protect\cite{dobbs2023smart} & 0.60 & 0.77 & 0.54 & \color{blue}0.71 \\
        Pc-Skeletor \protect\cite{molefe2023communication} & \color{blue}0.63 & \color{blue}0.79 &\color{blue} 0.56 & \color{blue}0.71 \\

        Ours & \color{red}\textbf{0.79} & \color{red}\textbf{0.84} & \color{red}\textbf{0.70} & \color{red}\textbf{0.78} \\
        \bottomrule
    \end{tabular}
    }
    \vspace{-1mm}
\caption{Performance comparison of different models on root phenotyping tasks. Evaluation metrics include Root Count Accuracy (RCA) and Longest Root Accuracy (LRA) for root systems with fewer than 100 roots (\textless{}100) and 500 roots (\textless{}500). The best results for each metric are highlighted in \textbf{\color{red}red}, and the second-best in \textbf{\color{blue}blue}.}
    \label{tab:performance_comparison}
    \vspace{-9mm}
\end{table}

\vspace{-2mm}
\subsection{ Comparison of 3D Root Skeleton Extraction}
\label{skeletoncompare}
\vspace{-1mm}
Currently, there is no widely accepted metric for evaluating 3D skeleton extraction specifically on plant root systems, as previous works have predominantly focused on skeletonization of objects or tree branches rather than roots. To demonstrate the effectiveness of our proposed method, we conducted a comparative study against state-of-the-art skeleton extraction algorithms, including methods from Wen et al. \cite{wen2023learnable}, Pc-Skeletor \cite{molefe2023communication}, Jiang et al. \cite{jiang2021skeleton}, AdTree \cite{du2019adtree}, and Smart-Tree \cite{dobbs2023smart}.
Figure~\ref{fig:pdf_root} shows a qualitative comparison of extracted skeletons. Our model better preserves fine, short roots and disentangles dense crossings. Wen et al.~\cite{wen2023learnable}, designed for everyday objects (e.g., chairs, cars), struggles with intricate biological root architectures. PC-Skeletor~\cite{molefe2023communication} applies Laplacian-based contraction with semantic priors (e.g., trunk/branch separation) that are unavailable or unreliable in roots; in dense regions, pure LBC often misconnects thin roots and forms spurious loops, contrary to biological expectations (diverging roots should not reconverge).

The methods from Jiang et al. \cite{jiang2021skeleton}, AdTree \cite{du2019adtree}, and Smart-Tree \cite{dobbs2023smart} are specifically tailored for extracting skeletons from tree point clouds. Due to the structural similarity between above-ground tree branches and root systems, these methods perform better than generic object-based algorithms. However, they still underperform compared to our approach, particularly in scenarios involving dense and complex root architectures.
For quantitative evaluation, we present a comparison in Table \ref{tab:performance_comparison}, where the extracted skeletons are assessed based on root count and the length of the longest root (i.e., the main root). Our model consistently outperforms the other methods across all metrics, demonstrating its effectiveness in accurately capturing detailed root skeletons, even in challenging cases with intricate root structures.
\begin{table}[t]
\vspace{-3mm}
    \centering
    \resizebox{0.48\textwidth}{!}{
    \begin{tabular}{lcccc}
        \toprule
        \textbf{Model} & \textbf{MC} & \textbf{DA} & \textbf{MC-F} & \textbf{DA-F} \\
                      & (\%) & (\%) & (\%) & (\%) \\
        \midrule
        ClipCap \cite{mokady2021clipcap} & 40.7 & 12.1 & 33.2 & 10.3 \\
        KRISP \cite{marino2021krisp} & 39.2 & 23.5 & 33.3 & 19.8 \\
        VLC-BERT \cite{zhou2020unified} & 48.4 & 34.5 & 40.1 & 31.0 \\
        GPV-2 \cite{kamath2022webly} & 51.3 & 37.2 & 43.0 & 34.1 \\
        Prophet-LLaMA \cite{shao2023prompting} & 58.6 & 42.5 & \color{blue}50.3 & 38.7 \\
        PromptCap+GPT-4o \cite{hu2023promptcap} & \color{blue}60.2 & \color{blue}51.3 & 49.9 & \color{blue}45.2 \\
        Ours & \color{red}\textbf{70.4} & \color{red}\textbf{59.7} & \color{red}\textbf{61.7} & \color{red}\textbf{50.3} \\
        \bottomrule
    \end{tabular}}
    \vspace{-1mm}
\caption{Performance comparison of different models on VQA tasks for plant root phenotyping.  MultipleChoice: MC, DirectAnswer: DA, MultipleChoice in few-shot questions: MC-F, and DirectAnswer in few-shot questions: DA-F. The best results for each metric are highlighted in \textbf{\color{red}red}, and the second-best in \textbf{\color{blue}blue}.}
    \label{tab:vqa_performance_comparison}
    \vspace{-9mm}
\end{table}
\subsection{Comparison of VQA Module Accuracy}
\vspace{-1mm}
To evaluate our fine-tuned GPT-4o on plant root VQA tasks, we compared it with multiple state-of-the-art VQA models. The comparison covered multiple-choice, true/false, and direct-answer questions. We also incorporated few-shot testing involving species or question types sparsely represented during training to assess generalization under limited data.
For models without built-in reasoning, we used GPT-4o to generate contextual inference, ensuring fair comparison by providing consistent contextual understanding and enabling better visual interpretation.
Figure~\ref{fig:pdf_image} presents representative test questions and outputs from our fine-tuned GPT-4o, spanning basic structural queries such as counts and types and context-dependent questions requiring visual--context integration. The model combines image evidence with metadata to produce coherent, accurate answers.
Quantitative results are summarized in Table~\ref{tab:vqa_performance_comparison}, reporting accuracy for multiple-choice and direct-answer questions, including few-shot scenarios. Our fine-tuned model with additional context consistently outperforms baselines across question types, with the largest gains in few-shot testing, where supplementary context improves generalization to unseen species and formats.
Overall, combining contextual data with fine-tuning significantly enhances performance in root phenotyping VQA. The strong generalization of fine-tuned GPT-4o underscores the effectiveness of leveraging contextual information and its inherent few-shot capability.
\begin{table}[t]
    \centering
    \resizebox{0.48\textwidth}{!}{
    \begin{tabular}{lcc}
        \toprule
        \textbf{Model} & \textbf{MC (\%)} & \textbf{DA (\%)} \\
        \midrule
        GPT-4o General Model \cite{achiam2023gpt} & 67.4 & 44.8 \\
        Model without 3D Skeleton Info & 66.7 & 43.5 \\
        Model using the same Skeleton Extraction & 68.3 & 47.2 \\
        Full Fine-Tuned Model & 70.4 & 59.7 \\
        \bottomrule
    \end{tabular}}
    \vspace{-1mm}
\caption{Performance comparison of different versions of our model on VQA tasks.  MultipleChoice: MC and DirectAnswer: DA. Results report the accuracy of the general GPT-4o model, the adaptive model without 3D skeleton extraction, the model using only the same skeleton extraction, and the full adaptive model with all components.}
    \label{tab:ablation_performance}
    \vspace{-10mm}
\end{table}

\subsection{Ablation Study}
\textbf{3D Skeleton Extraction}
The 3D skeleton extraction module provides quantitative context to GPT-4o during training and inference. Removing it eliminates these signals, degrading accuracy on attribute-specific questions and slightly reducing performance on related qualitative queries (e.g., effects of quantity on water absorption). As shown in Table~\ref{tab:ablation_performance}, this ablation causes a clear drop, particularly in detailed phenotypic analysis.
\textbf{ Skeletonization Losses}
We remove each loss term in turn. Without W-LBC, RCA/LRA drops substantially; without geometric consistency, both metrics remain lower than the full model; without linearity preservation, long-root accuracy degrades notably. The full model achieves the best RCA/LRA under both \textless{}100 and \textless{}500 settings (see Table~\ref{tab:ablation_performance_comparison}). This confirms that W-LBC governs contraction fidelity, while geometric and linearity constraints jointly stabilize topology and long-range continuity.
\begin{table}[!t]
\centering
\setlength{\tabcolsep}{3.5pt}
\resizebox{0.9\columnwidth}{!}{%
\begin{tabular}{lcccc}
    \toprule
    \textbf{Model}
    & \textbf{RCA}
    & \textbf{LRA}
    & \textbf{RCA}
    & \textbf{LRA} \\
    & \textbf{(\textless{}100)}
    & \textbf{(\textless{}100)}
    & \textbf{(\textless{}500)}
    & \textbf{(\textless{}500)} \\
    \midrule
    Without $\mathbf{L}_w$
    & 0.43 & 0.51 & 0.32 & 0.44 \\
    Without geometric loss
    & 0.72 & 0.73 & 0.62 & 0.66 \\
    Without linearity loss
    & 0.61 & 0.75 & 0.43 & 0.70 \\
    \midrule
    Full model
    & \textcolor{red}{\textbf{0.79}}
    & \textcolor{red}{\textbf{0.84}}
    & \textcolor{red}{\textbf{0.70}}
    & \textcolor{red}{\textbf{0.78}} \\
    \bottomrule
\end{tabular}%
}
\vspace{-1mm}
\caption{Performance comparison of our models without certain loss-relevant tasks. The metrics are the same as those in Table~\ref{tab:performance_comparison}.}
\label{tab:ablation_performance_comparison}
\vspace{-6mm}
\end{table}
\textbf{Adaption of GPT-4o Model}
In this experiment, we retained the 3D skeleton extraction and context generation modules but removed GPT-4o fine-tuning, instead using the general GPT-4o with the same inputs. Although the model still generates reasonable responses from the provided context, the lack of task-specific fine-tuning leads to a marked decrease in prediction accuracy. As shown in Table~\ref{tab:ablation_performance}, the accuracy of both multiple-choice and direct-answer questions drops significantly. Without fine-tuning, GPT-4o lacks the domain-specific adaptation required to interpret plant root phenotypic context effectively. This experiment highlights the role of fine-tuning in improving the model’s ability to answer complex domain-specific questions.
\begin{table}[t]
    \centering
    \begin{tabular}{lcccc}
        \toprule
         & \textbf{MC-FS} & \textbf{MC-ZS} & \textbf{DA-FS} & \textbf{DA-ZS} \\
                      & (\%) & (\%) & (\%) & (\%) \\
        \midrule
        Our Model & 61.7 & 53.2 & 50.3 & 33.4 \\
        \bottomrule
    \end{tabular}
    \vspace{-1mm}
\caption{Performance comparison of our model on VQA tasks under Few-shot (FS) and Zero-shot (ZS) settings. FS : species or question types with 1--5 labeled QA pairs seen during training; ZS : unseen species or question types at training.}
    \vspace{-9mm}
    \label{tab:few_shot_comparison}
\end{table}
\textbf{Few-Shot Learning.}
We evaluate few-shot learning by comparing datasets with and without designated low-shot samples. In the few-shot setting, 15\% of the training data contains only 1--5 labeled question--answer pairs per species or question type, while the remaining 85\% follows standard supervised training. As shown in Table~\ref{tab:few_shot_comparison}, including these samples improves accuracy on rare species and complex questions, whereas removing them substantially degrades performance on unseen species and novel question types. These results demonstrate that few-shot supervision enhances GPT-4o's generalization to underrepresented phenotyping scenarios.
\vspace{-2mm}
\section{Conclusion}
We propose a multimodal framework for plant root phenotyping that combines unsupervised 3D skeleton extraction with evidence-grounded language reasoning. The framework converts dense root point clouds into structured representations and uses a fine-tuned GPT-4o model to generate interpretable phenotypic analysis. Results show improved structural accuracy, reasoning quality, and generalization.

 \textbf{Acknowledgment}: This work is supported by USDA NIFA grant No. 2021-67021-34199 and NSF Grants NO. 2340882, 2334624, 2334246, and 2334690.
{\small
\bibliographystyle{ieee_fullname}
\bibliography{main}
}

\end{document}